\documentclass[sigconf]{acmart}
\AtBeginDocument{%
  }

\copyrightyear{2025}
\acmYear{2025}
\setcopyright{cc}
\setcctype{by}
\acmConference[SIGIR '25]{Proceedings of the 48th International ACM SIGIR Conference on Research and Development in Information Retrieval}{July 13--18, 2025}{Padua, Italy}
\acmBooktitle{Proceedings of the 48th International ACM SIGIR Conference on Research and Development in Information Retrieval (SIGIR '25), July 13--18, 2025, Padua, Italy}
\acmDOI{10.1145/3726302.3730070}
\acmISBN{979-8-4007-1592-1/2025/07}

\usepackage{balance}

\begin{document}

\title{Reason-to-Rank: Distilling Direct and Comparative Reasoning from Large Language Models for Document Reranking}

\author{Yuelyu Ji}
\orcid{0000-0001-6389-5823}  
\affiliation{%
  \institution{University of Pittsburgh}
  \city{Pittsburgh}
  \state{PA}
  \country{USA}
}
\email{yuj49@pitt.edu}

\author{Zhuochun Li}
\orcid{0009-0004-0772-9416}
\affiliation{%
  \institution{University of Pittsburgh}
  \city{Pittsburgh}
  \state{PA}
  \country{USA}
}
\email{zhl163@pitt.edu} 

\author{Rui Meng}
\orcid{0000-0001-5583-4924}
\affiliation{%
  \institution{Google Cloud AI Research}
  \city{Sunnyvale}
  \state{CA}
  \country{USA}
}
\email{rmeng@google.com} 

\author{Daqing He}
\orcid{0000-0002-4645-8696}
\affiliation{%
  \institution{University of Pittsburgh}
  \city{Pittsburgh}
  \state{PA}
  \country{USA}
}
\email{dah44@pitt.edu}

\begin{abstract}

 Reranking documents in information retrieval often relies on black-box models that improve effectiveness but lack explainability. We introduce Reason-to-Rank (R2R), a novel framework that separates direct relevance reasoning from comparison reasoning to provide both direct and comparitive explanations. We first prompt a large language model to produce comprehensive rationales and a ranking order; then we distill both the ranking decisions and textual explanations into a smaller, open-source student model. Our approach not only improves retrieval performance, as demonstrated in MSMARCO, BEIR, and BRIGHT, but also provides interpretable justifications for why one document outranks another. We report NDCG@5 (and NDCG@10) for direct comparisons with prior work, and show that the distilled student model achieves competitive results while significantly reducing computational overhead. By unifying direct and comparative reasoning in a single pipeline, R2R bridges the gap between transparency and effectiveness in modern reranking systems.
\end{abstract}



\maketitle

\section{Introduction}
\label{sec:intro}
\begin{figure}[t]
  \centering
  \includegraphics[width=0.45\textwidth]{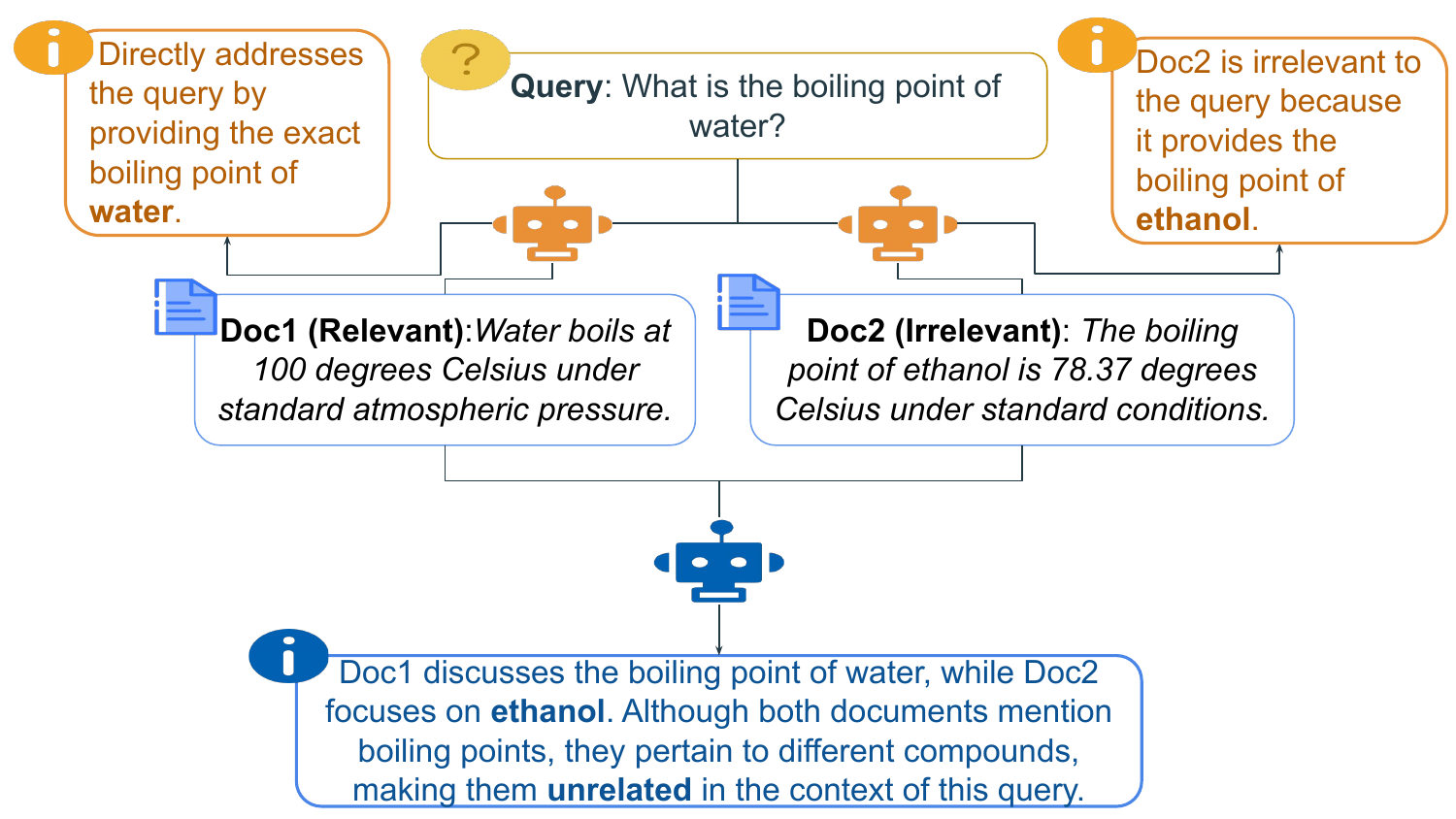}
  \caption{Illustration of the two types of reasoning: \textcolor{orange}{ direct relevance reasoning} provides explicit answers to the query, while \textcolor{blue}{comparison reasoning} evaluates the relative relevance between documents. The LLM generates these explanations to enhance the interpretability of the ranking process.}
  \label{fig:reasoning}
\end{figure}
With the explosive growth of online information, modern search engines and recommendation systems heavily rely on \emph{reranking} to refine initial retrieval results and present users with the most relevant content~\cite{sun2023chatgpt,pradeep2023rankzephyr}. Despite significant progress in neural reranking, particularly with large language models (LLMs) such as GPT-4, contemporary solutions often lack transparency in how they differentiate between highly similar candidate documents. Users see a final ranking list, yet remain uncertain about \emph{why} how specific items are prioritized, which can reduce trust and hinder effective system debugging~\cite{nogueira2020document,liu2024pe_rank}.

\textbf{Explainable reranking} has attracted increasing attention, with the aim of combining ranking decisions with human-readable justifications. Recent studies~\cite{yu2024explain,zhuang2023beyond} show that providing textual explanations can improve user understanding and confidence, and can also help system designers pinpoint potential errors in the ranking pipeline. However, most prior approaches focus on a single dimension of explanation, \emph{direct relevance reasoning}, where each document is described in isolation: ``This document is relevant to the query because of...''. While useful, such \emph{direct} rationales do not address the \emph{relative} importance of documents, a critical aspect when multiple items appear similarly relevant yet differ in subtle ways.

We argue that \emph{comparison reasoning} is the missing piece in many existing explainable reranking systems. By comparing pairs (or subsets) of documents---explaining, for instance, \emph{why Document A is placed above Document B}---the model can highlight unique advantages or deficiencies, enabling more fine-grained control over the final ranked list~\cite{pradeep2021expando,sun2023chatgpt}. Without such pairwise or listwise clarity, users and IR practitioners must rely on guesswork when interpreting top results, limiting the overall efficacy of explanation-based retrieval solutions.

In this paper, we propose \textbf{Reason-to-Rank (R2R)}, a novel framework that unifies \emph{direct relevance} with \emph{comparison-based} reasoning. R2R leverages a powerful \emph{teacher} LLM (e.g., GPT-4) to produce both a global ranking and two kinds of textual justifications: (1)~a direct explanation for each candidate, and (2)~a comparative explanation indicating which document outranks another and why. The framework is shown in Figure \ref{fig:reasoning}.We then employ a \emph{knowledge distillation} procedure to train a smaller \emph{student} model to replicate the teacher's ranking decisions and explanations. From an educational perspective, research shows that learners benefit more when they are taught not just answers but the reasoning behind them~\cite{raza2020intelligent}. Inspired by this, our approach transfers both the \emph{what} and the \emph{why}, helping the student model generalize better and generate its own explanations with improved clarity.
 This setup offers two major advantages:  
\emph{(i)}~The final deployed system preserves interpretability without incurring the high computational cost of querying a large LLM in production, and  
\emph{(ii)}~By explicitly encoding comparative rationales, R2R achieves more subtlety distinctions among the top candidates than direct relevance only methods.

We perform extensive experiments on multiple benchmarks, including \textbf{MSMARCO}~\cite{bajaj2016ms}, \textbf{BEIR}~\cite{thakur2021beir}, and \textbf{BRIGHT}~\cite{su2024bright}, covering both general search queries and reasoning-intensive tasks. The results show that R2R outperforms various baselines (e.g., monoT5~\cite{nogueira2020document}, RankVicuna~\cite{pradeep2023rankvicuna}) in terms of NDCG metrics, while also providing more interpretable output. A comprehensive ablation study further validates the importance of integrating both direct relevance and comparison reasoning, as omitting either dimension leads to decreased performance and reduced clarity. In addition, we provide qualitative examples illustrating how comparison explanations help reveal subtle but important differences among retrieved documents.

In summary, our primary contributions are:
\begin{itemize}
    \item We introduce a novel reranking framework, \textbf{R2R}, that produces both \emph{direct relevance} and \emph{comparison-based} explanations, offering a more transparent and fine-grained view of document ranking.
    \item We propose a \emph{teacher--student} knowledge distillation pipeline, where a large LLM generates enriched multiperspective rationales, subsequently transferred to a smaller, more efficient student model for practical deployment.
    \item We conduct extensive evaluations on general and reasoning-focused datasets, demonstrating that R2R achieves not only competitive or superior reranking performance but also enhanced interpretability through dual-type reasoning.
\end{itemize}

\begin{figure*}[t]
  \centering
  \includegraphics[width=\textwidth]{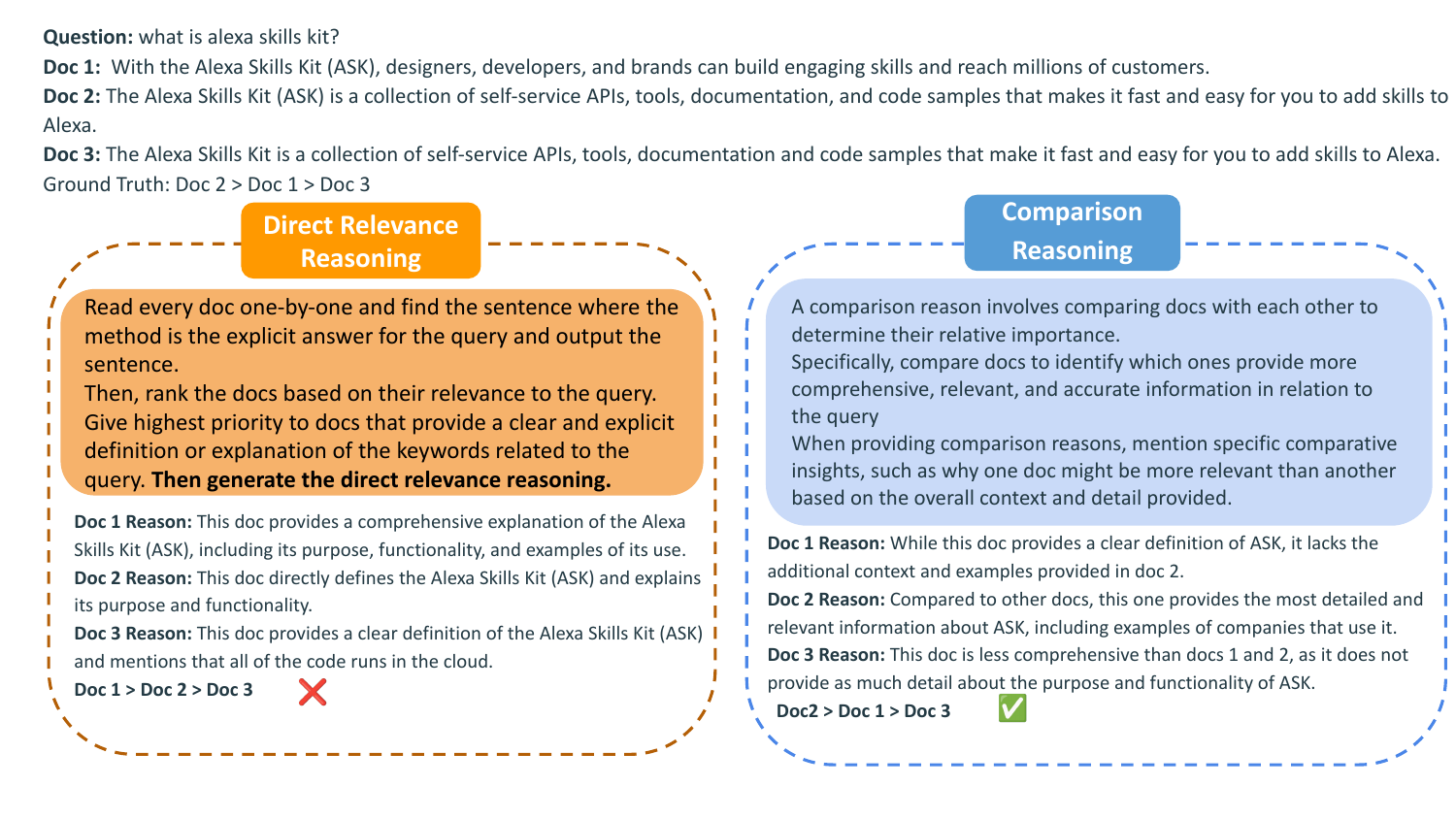}
  \caption{Overview of  direct relevance and comparison reasoning prompts for document ranking. Direct relevance reasoning explains how a document matches a query, while comparison reasoning evaluates the relative relevance between documents.}
  \label{fig:prompts}
\end{figure*}
\section{Related Work}
\label{sec:related_work}
\subsection{Neural Reranking and Explanation}
Neural reranking has been extensively studied to refine candidate lists obtained from first-stage retrieval (e.g., BM25 or dense retrievers)~\cite{nogueira2019multi,nogueira2020document}. Early efforts often used cross-encoder architectures (e.g., BERT or T5) to capture fine-grained query--document interactions. Although these models achieve strong effectiveness, they typically present limited insights regarding the \emph{reasoning} behind their decisions~\cite{liu2009learning}. 

Recent work has thus explored generating \emph{explanations} alongside the ranking~\cite{zhuang2023beyond,yu2024explain}. For instance, Explain-then-Rank~\cite{zhuang2023beyond} adds textual justifications on how a document addresses the query. However, these methods tend to focus on \emph{direct relevance} (i.e., how well a single document matches the query) rather than explicitly modeling \emph{comparative} relevance among documents, which may be crucial in distinguishing subtle differences at the top of the ranked list~\cite{doshi2017towards,lipton2018mythos}.

\subsection{Direct vs.\ Comparison Reasoning}
Classical Learning to Rank (LTR) frameworks~\cite{liu2009learning} have shown that pointwise, pairwise, and listwise strategies each possess unique advantages. Pairwise approaches like duoBERT~\cite{pradeep2021expando} are explicitly designed to compare two documents and decide which one is more relevant, whereas pointwise methods focus on rating each document in isolation. More recently, listwise approaches (e.g., RankGPT~\cite{sun2023chatgpt} or RankVicuna~\cite{pradeep2023rankvicuna}) consider the entire ranked list, often achieving stronger empirical results. Nonetheless, even these advanced listwise techniques do not always clearly \emph{explain} why one document outranks another. Our proposed R2R differs by \emph{explicitly} producing both direct relevance and comparative rationales within a unified framework.

\subsection{Knowledge Distillation in Reranking}
Recent work has addressed the dual goals of ranking \emph{effectiveness} and \emph{transparency}. On the explanation front, \textbf{EXS}~\cite{singh2019exs} offers local, model-agnostic rationale extraction for pointwise rankers, whereas \textbf{RankingSHAP}~\cite{heuss2024rankingshap} applies Shapley-value analysis across entire ranked lists. Both methods clarify \emph{why} documents are preferred, but do not compress these insights into smaller, more efficient models.

Meanwhile, methods like \textbf{One-Shot Labeling}~\cite{macavaney2023one} and \textbf{Few-Shot Prompting for Pairwise Ranking}~\cite{sinhababu-etal-2024-shot} demonstrate that even sparse or minimal relevance data can significantly boost retrieval. Although they refine ranking accuracy via limited supervision, they typically do not target explanation generation or model compression.

Overall, our approach is complementary: we incorporate both direct and comparative reasoning, and distill these insights into an open-source student model that can both rank and explain. Extending Reason-to-Rank to incorporate advanced explanation techniques such as Shapley-based analysis is a promising avenue for future work.

\begin{figure}[ht]
\centering
\includegraphics[width=0.45\textwidth]{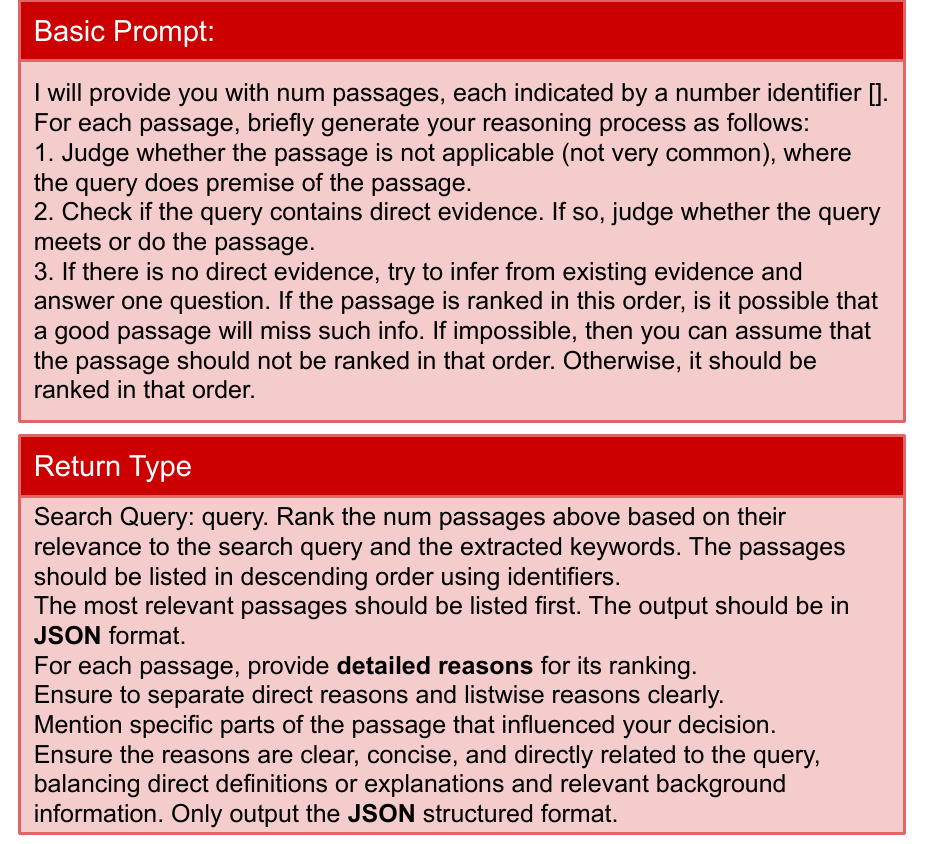}
\caption{Basic prompt and return type.}
\label{fig:basic_prompt}
\end{figure}

\begin{figure*}[t]
  \centering
  \includegraphics[width=\textwidth]{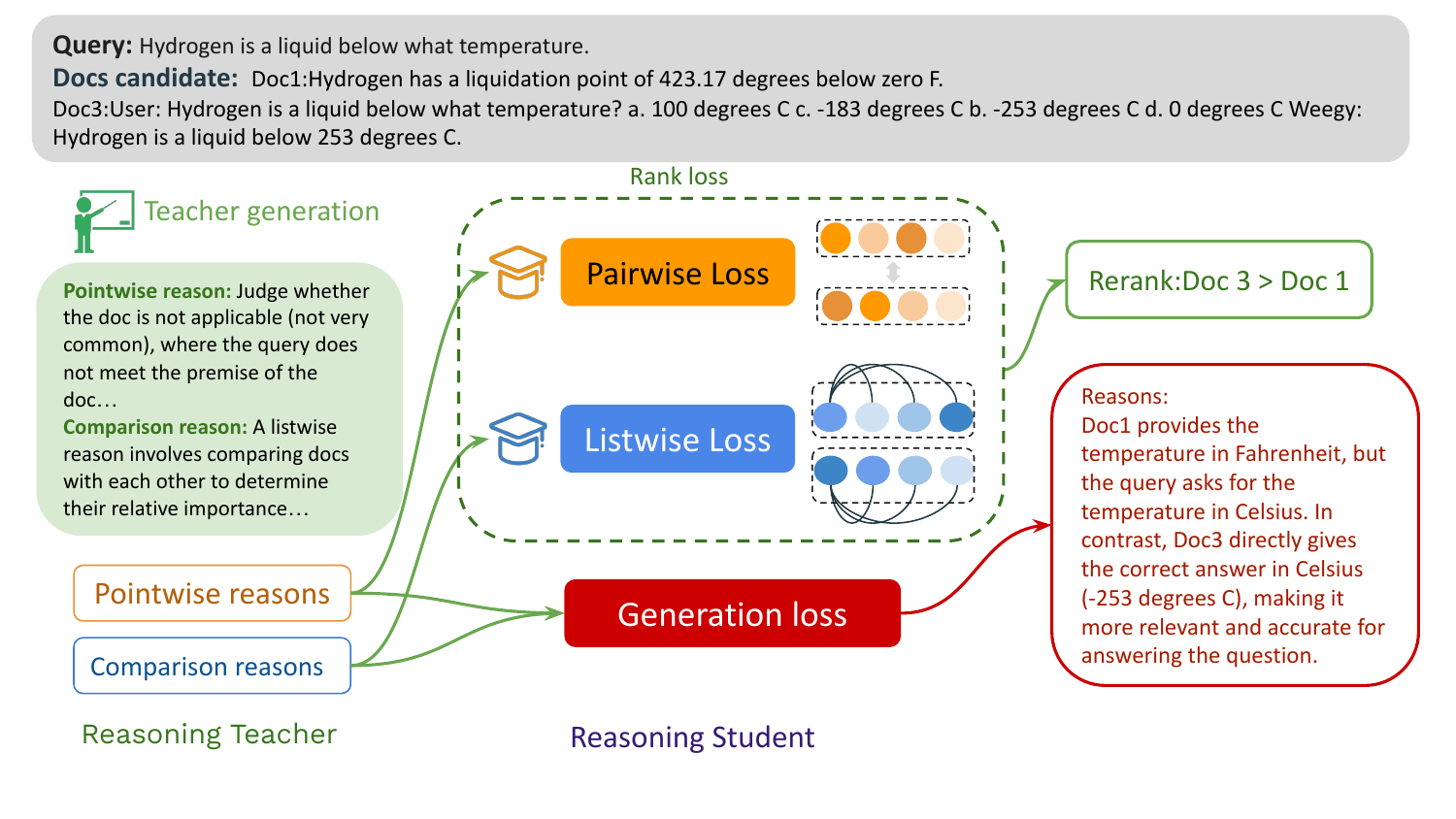}
  \caption{Overview of the Reason-to-Rank framework. The teacher model generates  direct relevance and comparative reasoning, which is used to train student models capable of reproducing the reranked order and generating explanations.}
  \label{fig:framework}
\end{figure*}

\section{Method}
\label{sec:methodology}
 
\paragraph{Initial Retrieval with BM25.}
Following common practice in neural re-ranking~\cite{nogueira2019multi,sun2023chatgpt}, we first retrieve the top-100 candidate documents per query using BM25. This initial retrieval step serves as the candidate pool from which our re-ranking pipeline will operate. We choose 100 to balance recall (ensuring that most relevant documents are included) and computational overhead.\footnote{Our code and data are available here \url{https://github.com/JoyDajunSpaceCraft/Reason-to-Rank.git}}

\subsection{Teacher Model with Dual Reasoning}
\label{sec:teacher_model}
Given a query \( q \) and a set of candidate documents \( \{d_1, d_2, \dots, d_n\} \), the teacher model processes these inputs to generate two types of reasoning and rerank order.
The teacher model can generate high-quality reasoning that combines both  direct relevance and comparative aspects, and the examples can be seen in Figure \ref{fig:prompts}:
\textbf{ Direct relevance Reasoning}: The model explains how each document directly addresses the query, focusing on the relevance and specificity of the content.
\textbf{Comparative Reasoning}: The model assesses the relative relevance between documents, explaining why one document should be ranked higher based on content. 
We also have the basic prompt and the return type in Figure \ref{fig:basic_prompt}.
Our teacher model (e.g., GPT-4) processes the top-100 candidates and produces two outputs:
\begin{enumerate}
    \item A \emph{ranking order} (i.e., a global ordering of the 100 documents), which corresponds to how the teacher prioritizes each document’s relevance to the query.
    \item A \emph{textual explanation} for each document, covering both direct relevance (i.e., how the document addresses the query) and comparison reasoning (i.e., why one document is more relevant than another).
\end{enumerate}
While the teacher internally may compute or utilize relevance \emph{scores}, we unify these into a final \emph{ordering} to avoid ambiguity. In other words, rather than exposing raw numeric scores, the teacher model outputs a definitive rank list (\textit{e.g.}, from most relevant to least relevant) alongside the corresponding rationales.

\subsection{Student Model with Knowledge Distillation}
Deploying large language models (LLMs) for reranking is often impractical due to their high computational cost. To address this, we design the student model to replicate the teacher's performance while being more efficient for real-world deployment. We obtain numeric ranking scores for supervision by prompting the teacher to produce a ranked list and converting the document positions into scores via normalized ranks. The innovation lies in using knowledge distillation: the student learns from the teacher’s reasoning and ranking outputs, retaining accuracy and interpretability with significantly lower computational demands. This approach ensures the student can perform complex reranking tasks efficiently, making it ideal for large-scale use.

\subsubsection{Model Overview}
Given a query \( q \) and a set of candidate documents \( \{d_1, d_2, \dots, d_n\} \), the model processes these inputs to generate both reasoning outputs and ranking scores.

The model generates two outputs:
\textbf{Reasoning Outputs:} The model generates textual reasons \( \mathbf{r}_i \) explaining the relevance of each document \( d_i \) to the query \( q \).
\textbf{Ranking Scores:} The model simultaneously generates ranking scores \( \mathbf{s}_i \) for each document, indicating its relevance to the query.

\subsubsection{Training Objective}

To optimize the performance of the student model, we define three distinct loss functions, each corresponding to different aspects of the model. First, the pairwise loss optimizes the relative ranking between pairs of documents, ensuring that the more relevant document is ranked higher. The pairwise loss is defined as:

\[
\mathcal{L}_{\text{pairwise}} = \sum_{(i,j) \in P} \max(0, 1 - (\mathbf{s}_i - \mathbf{s}_j))
\]

where \( \mathbf{s}_i \) and \( \mathbf{s}_j \) represent the predicted relevance scores of documents \( d_i \) and \( d_j \), and \( P \) is the set of document pairs with known relevance relations.

Next, the listwise loss optimizes the overall ranking of the document list. It uses Kullback-Leibler (KL) divergence \cite{hershey2007approximating} to measure the difference between the predicted ranking distribution and the target ranking distribution:

\[
\mathcal{L}_{\text{listwise}} = D_{\text{KL}}(Q(\mathbf{z}) \parallel P(\mathbf{s}))
\]

where \( P(\mathbf{s}) = \text{softmax}(\mathbf{s}_i) \) is the predicted ranking distribution based on the model’s output, and \( Q(\mathbf{z}) = \text{softmax}(\mathbf{z}_i) \) is the target ranking distribution based on true labels.
We treated the pairwise loss and listwise loss as rank loss.
Finally, the generation loss optimizes the quality of the textual reasoning generated by the model. It is based on the cross-entropy loss between the generated reasoning and the ground-truth reasoning:

\[
\mathcal{L}_{\text{generation}} = -\sum_{i=1}^{n} \mathbf{y}_i^{\text{gen}} \log(\text{softmax}(\mathbf{r}_i))
\]

where \( \mathbf{r}_i \) represents the generated reasoning logits for document \( d_i \), and \( \mathbf{y}_i^{\text{gen}} \) is the ground-truth reasoning label.

To achieve balanced training, we introduce learnable weights \( \alpha \), \( \beta \), and \( \gamma \) to adjust the contributions of pairwise, listwise, and generation losses, respectively. The final loss function is:

\[
\mathcal{L} = \alpha \times \mathcal{L}_{\text{pairwise}} + \beta \times \mathcal{L}_{\text{listwise}} + \gamma \times \mathcal{L}_{\text{generation}}
\]

where \( \alpha + \beta + \gamma = 1 \), ensuring a balanced combination of the three loss functions during training. With this design, the student model efficiently performs ranking tasks and provides transparent reasoning for its decisions, enhancing both transparency and performance.

The overview of the architecture of the R2R model is shown in Figure \ref{fig:framework}. \textbf{R2R}, a framework that integrates direct relevance and comparative reasoning within a unified reranking approach.


\begin{table*}[ht]
\centering

\caption{NDCG@5 performance (in percentage) for different models with various reasoning prompts across multiple datasets. \textbf{Bold} numbers indicate the best performance within each group of models. $^{*}$ indicates statistically significant improvement over the base model without reasoning prompts (p $<$ 0.05). $^{\dagger}$ indicates statistically significant improvement over both the base model and the single reasoning prompt variants (p $<$ 0.05).DL19 and DL20 \cite{craswell2020overview} are test datasets for MSMARCO. The MonoT5 has no result for News, and NFCorpus and RankVicuna has no results for News in their paper.}

\small
\setlength{\tabcolsep}{4pt} 
\renewcommand{\arraystretch}{1.1} 
\resizebox{\textwidth}{!}{
\begin{tabular}{lccc|cccccc}
\toprule
\textbf{Models} & \textbf{DL19} & \textbf{DL20} & \textbf{BEIR-6 Avg.} & \textbf{Covid} & \textbf{Touche} & \textbf{News} & \textbf{NFCorpus} & \textbf{Robust04} & \textbf{DBPedia} \\ \midrule

BM25 & 50.6 & 48.0 & 41.0 & 59.5 & 44.2 & 39.5 & 30.8 & 40.7 & 31.8 \\ \midrule

\multicolumn{10}{c}{\textbf{Rank GPT-4 Model}}\\ \midrule
Rank GPT-4 (w/o reasoning) & 75.6 & 70.6 & 54.0 & 85.5 & 38.5 & 57.6 & 36.6 & 57.6 & 47.1 \\
\quad + direct relevance  reasoning & 75.3$^{*}$ & 69.9$^{*}$ & 52.9 & 83.3$^{*}$ & \textbf{38.9} & 55.9$^{*}$ & 36.1 & 52.9$^{*}$ & 47.0 \\ 
\quad + comparison reasoning & 76.3$^{*}$ & 71.1$^{*}$ & 53.4 & 84.6$^{*}$ & 38.2 & 56.5$^{*}$ & 36.5 & 56.5$^{*}$ & 48.2$^{*}$ \\ 
\quad +  direct relevance  \& comparison & \textbf{77.7}$^{\dagger}$ & \textbf{73.2}$^{\dagger}$ & \textbf{54.4} & \textbf{85.3}$^{\dagger}$ & 38.3 & \textbf{58.4}$^{\dagger}$ & 36.3 & \textbf{58.6}$^{\dagger}$ & \textbf{49.5}$^{\dagger}$ \\ \midrule

\multicolumn{10}{c}{\textbf{Claude Model}}\\ \midrule
Claude (w/o reasoning) & 72.0 & 68.8 & 50.8 & 82.8 & 36.1 & 54.4 & 36.4 & 50.8 & 42.5 \\ 
\quad + direct relevance  reasoning & 70.6$^{*}$ & 68.2 & 50.3 & 82.2 & 35.9 & 55.4$^{*}$ & 35.6 & 52.4$^{*}$ & 42.5 \\ 
\quad + comparison reasoning & 71.3 & 69.8$^{*}$ & 51.1 & 83.3$^{*}$ & 36.4 & 55.1$^{*}$ & 36.7 & 51.5 & 43.2$^{*}$ \\ 
\quad +  direct relevance \& comparison & 72.1$^{\dagger}$ & 70.0$^{\dagger}$ & 52.0 & 84.0$^{\dagger}$ & 37.3$^{\dagger}$ & 55.5$^{\dagger}$ & 36.4 & 52.7$^{\dagger}$ & 44.9$^{\dagger}$ \\ \midrule

\multicolumn{10}{c}{\textbf{Gemini Model}}\\ \midrule
Gemini (w/o reasoning) & 71.1 & 68.5 & 50.1 & 83.0 & 35.6 & 53.7 & 35.5 & 50.1 & 42.9 \\ 
\quad + direct relevance  reasoning & 69.1$^{*}$ & 67.0$^{*}$ & 51.1 & 81.6$^{*}$ & 34.8 & 52.9$^{*}$ & \textbf{36.9}$^{*}$ & 51.3$^{*}$ & 43.9$^{*}$  \\ 
\quad + comparison reasoning & 70.2$^{*}$ & 68.2$^{*}$ & 50.2 & 82.4$^{*}$ & 35.9 & 53.4 & 35.4 & 51.1$^{*}$ & 42.9 \\ 
\quad +  direct relevance \& comparison & 71.4$^{\dagger}$ & 68.6$^{\dagger}$ & 51.0 & 83.5$^{\dagger}$ & 37.0$^{\dagger}$ & 53.8$^{\dagger}$ & 36.1 & 51.8$^{\dagger}$ & 43.7$^{\dagger}$ \\ 
\bottomrule
\end{tabular}}
\label{tab:ndcg5_teachers}
\end{table*}
\begin{table*}[ht]
\centering

\caption{NDCG@5 performance (in percentage) for student models and baseline comparisons across multiple datasets. 'w/o Reasoning' models do not use reasoning prompts and do not generate explanations; they are optimized using the specified ranking loss functions. 'w/  direct relevance Reasoning' and 'w/ comparison Reasoning' models use reasoning prompts and include reasoning loss for generating explanations. '$^{*}$' indicates statistically significant improvement over the baseline DeBERTa model (p $<$ 0.05). \textbf{Bold} indicates the best performance among all models for each dataset. Entries marked with '--' indicate that results are not available for that dataset.}

\small
\setlength{\tabcolsep}{4pt}
\renewcommand{\arraystretch}{1.15}
\resizebox{\textwidth}{!}{
\begin{tabular}{lccc|cccccc}
\toprule
\textbf{Models} & \textbf{DL19} & \textbf{DL20} & \textbf{BEIR-6 Avg.} & \textbf{Covid} & \textbf{Touche} & \textbf{News} & \textbf{NFCorpus} & \textbf{Robust04} & \textbf{DBPedia} \\
\midrule

\multicolumn{10}{c}{\textbf{Baseline Models}} \\ \midrule

DeBERTa \cite{sun2023chatgpt} & 68.5 & 64.2 & 47.0 & 73.4 & 32.1 & 50.2 & 33.7 & 49.2 & 45.4 \\
MonoT5 \cite{nogueira2020document} & 74.5 & 70.4 & 48.3 & 80.0 & 34.1 & -- & -- & 46.0 & 35.2 \\ 
RankVicuna \cite{pradeep2023rankvicuna} & 71.1 & 68.7 & 48.3 & 67.1 & \textbf{48.7} & -- & 38.5 & \textbf{55.7} & 35.3 \\ 
RankZephyr \cite{pradeep2023rankzephyr} & 72.2 & 70.5 & 51.3 & \textbf{85.1} & 36.5 & \textbf{53.3} & 38.9 & \textbf{60.7} & 35.5 \\ 
APEER \cite{jin2024apeer} & \textbf{74.6} & \textbf{72.3} & 51.1 & 83.9 & 35.3 & 52.1 & 33.4 & 56.0 & 46.1\\
\midrule

\multicolumn{10}{c}{\textbf{Our Models (student)}} \\ \midrule
w/o reasoning (Pairwise Loss) & 73.2 & 70.8 & 50.1 & 79.9 & 35.1 & 51.9 & 34.6 & 52.8 & 46.3 \\
w/o  reasoning (Listwise Loss) & 73.8 & 71.0 & 50.5 & 80.2 & 35.4 & 52.3 & 35.0 & 53.0 & 46.6 \\
w/  direct relevance reasoning & 74.5$^{*}$ & 70.1 & 49.8 & 79.4 & 34.3 & 52.8$^{*}$ & 35.5$^{*}$ & 50.8 & 47.0 \\
w/ comparison reasoning& 74.1$^{*}$ & 72.3$^{*}$ & 50.9 & 80.1$^{*}$ & 34.0 & 53.1$^{*}$ & 36.6$^{*}$ & 52.1 & 47.2$^{*}$ \\
w/  direct relevance \& comparison & \textbf{75.4}$^{*}$ & \textbf{72.4}$^{*}$ &\textbf{ 52.4 }$^{*}$& 84.6$^{*}$ & 36.2$^{*}$ & 53.8$^{*}$ & 36.4$^{*}$ & 53.5$^{*}$ & \textbf{47.9}$^{*}$ \\
\bottomrule
\end{tabular}
}

\label{tab:ndcg5_student_baseline}
\end{table*}

\section{Experiments}
\label{sec:experiments}
In this section, we evaluate the performance of our \textbf{Reason-to-Rank} (R2R) framework across various information retrieval tasks. We introduce the datasets and evaluation metrics used in our experiments \S\ref{sec:datasets}. We then detail the experimental setup and baseline methods for comparison \S\ref{sec:setup}. Finally, we present a comprehensive analysis of the experimental results, including comparisons with existing methods in \S\ref{sec:comparative} and \S\ref{teacer_mode}.

\subsection{Datasets and Evaluation Metrics}
\label{sec:datasets}
In addition to MSMARCO \cite{bajaj2016ms} and BEIR \cite{thakur2021beir}, we utilize the BRIGHT \cite{su2024bright} to further evaluate our model’s ability to handle reasoning-intensive retrieval tasks. 
\paragraph{Evaluation Metrics}
We use the following evaluation metrics to measure the performance of our models:

 \textbf{Normalized Discounted Cumulative Gain (NDCG@k)}: Measures the ranking quality of the retrieved documents, with higher emphasis on top-ranked documents. We report NDCG at rank positions 5 and 10.
 
\textbf{BLEU}, and \textbf{ROUGE-L}: Used for evaluating the quality of generated reasons in reasoning tasks, assessing both lexical overlap and logical consistency.

\subsection{Models}
\label{sec:setup}

\subsubsection{Teacher models}

We utilize large language models (LLMs) as teacher models to generate reasoning annotations: \textbf{GPT-4 \cite{achiam2023gpt}}, \textbf{Claude \cite{TheC3}} (claude-3-5-sonnet-20240620) and  \textbf{Gemini \cite{team2023gemini}} (gemini-1.5-flash). These advanced LLMs known for their strong reasoning and language-understanding capabilities.

\paragraph{Teacher Model Performance}
\label{teacer_mode}
Table~\ref{tab:ndcg5_teachers} summarizes the NDCG@5 performance of different models \textbf{GPT-4} consistently achieves the best results, with \textbf{77.6\%} on DL19 and \textbf{85.2\%} on TREC-COVID when using both comparison and direct relevance reasoning prompts, showcasing its robust handling of complex queries.

Adding different reasoning prompts to the LLMs enhances their performance across various datasets. For instance, GPT-4 with comparison and direct relevance reasoning outperforms its baseline (without reasoning) on DL19 by \textbf{2.1\%}. Similar trends are observesd with Claude and Gemini, indicating that incorporating reasoning prompts in different LLMs is beneficial.
To confirm the statistical significance of the improvements, we conduct paired t-tests comparing models with reasoning prompts to their respective baselines.  Given GPT-4's overall solid performance and the boost from reasoning prompts, we select it as the teacher model to distill its enhanced reasoning abilities into our student models.

\subsubsection{Student models}

We implement our models using LoRA (Low-Rank Adaptation) on 32GB V100 GPUs to fine-tune the student models on large datasets efficiently. Based on the \textbf{LLaMA 3.1 8B} architecture, the student models fine-tuned with different reasoning prompts derived from the teacher models.

\paragraph{Impact of Reasoning Strategies}
we also evaluated different reasoning strategies in the R2R student models, specifically analyzing the "w/o Reasoning" (without reasoning), "w/  direct relevance reasoning" (with  direct relevance reasoning), "w/ comparison reasoning" (with comparison reasoning), and "w/  direct relevance \& comparison" (with both  direct relevance and comparison reasoning) variants.

As shown in Table ~\ref{tab:ndcg5_student_baseline},  models without reasoning (using pairwise or listwise loss) perform at a baseline level, with NDCG@5 scores of \textbf{73.2\%} and 73.8\% on DL19 and \textbf{70.8\%} and \textbf{71.0\% }on DL20. Adding reasoning significantly boosts performance. For example,  direct relevance reasoning improves NDCG@5 to \textbf{74.5\%} on DL19 and \textbf{70.1\%} on DL20 by providing  direct relevance relevance explanations. Comparison reasoning, which assesses the relative importance between documents, raises NDCG@5 to \textbf{74.1\%} on DL19 and \textbf{72.3\%} on DL20. Combining both types of reasoning yields the best results, with NDCG@5 scores of \textbf{75.4\%} on DL19 and \textbf{72.4\%} on DL20.

Interestingly, while reasoning improves performance in most cases, there are instances where models without reasoning (e.g., listwise loss) outperform those with reasoning. 

\subsubsection{Comparison with Baselines}
Our student models, trained with  direct relevance and comparison reasoning, achieve competitive performance compared to the previous state-of-the-art student models. As shown in Table ~\ref{tab:ndcg5_student_baseline},the improvements over the baseline models are statistically significant, with p-values less than \textbf{0.05}.

\textbf{BM25}: A traditional term-based retrieval model using TF-IDF, known for its simplicity and efficiency in keyword-based retrieval.

\textbf{RankGPT DeBERTa} \cite{sun2023chatgpt}: A transformer model using disentangled attention to better capture long-range dependencies, improving reranking by understanding query-document context.

\textbf{MonoT5} \cite{nogueira2020document}: A T5-based model treating reranking as a sequence generation task, leveraging its encoder-decoder architecture for better relevance scoring.

\textbf{RankVicuna}, \textbf{RankZephyr} and \textbf{APEER} 
\cite{pradeep2023rankvicuna,pradeep2023rankzephyr,jin2024apeer}: 
\textbf{RankVicuna} is a distilled version of RankGPT, designed for zero-shot listwise reranking with fewer parameters, balancing efficiency and performance. \textbf{RankZephyr} a state-of-the-art zero-shot listwise reranking model that adapts RankGPT prompts for robust performance with minimal task-specific training. \textbf{APEER} share the same structure with RankZephyr but have an automated prompt generation for large language models. 

\subsubsection{R2R Student Model vs. Baselines}
\label{sec:comparative}
Our R2R student model indicates strong performance across various datasets, often outperforming the baseline models. For instance, on the DL19 dataset, our student model achieves an NDCG@5 of \textbf{75.3\%}, surpassing DeBERTa (\textbf{68.5\%}), RankVicuna (\textbf{71.1\%}), and RankZephyr (\textbf{72.2\%}). Similarly, on the DL20 dataset, our model attains \textbf{72.3\%}, outperforming DeBERTa (\textbf{64.1\%}) and RankVicuna (\textbf{68.6\%}).

On some datasets, our model achieves the best performance among all models. For example, on the \textbf{News} dataset, our student model reaches an NDCG@5 of \textbf{53.8\%}, slightly outperforming RankZephyr (\textbf{53.3\%}) and significantly better than DeBERTa (\textbf{50.2\%}).

However, there are cases where other models perform better. On the \textbf{Touche} dataset, RankVicuna achieves an NDCG@5 of \textbf{48.7\%}, which is higher than our student model's \textbf{36.2\%}. This may be due to RankVicuna's design for zero-shot listwise reranking, which could be particularly effective for argument retrieval tasks represent in Touche. Similarly, on the \textbf{Robust04} dataset, RankZephyr attains a higher NDCG@5 of \textbf{60.7\%} compared to our model's \textbf{53.5\%}.

For Touche, it is shown that it is a \textbf{argument retrieval} or \textbf{debate scenario} dataset, and BM25 tends to perform surprisingly well in such “keyword-matching” strong correlation scenarios; furthermore, some models (e.g., RankVicuna) may be better adapted in argument mining Furthermore, some models (e.g., RankVicuna) may be better adapted for argument mining.
For Robust04, this means that it covers a wider range of topics and outdated documents, where RankZephyr's multi-iteration listwise training is more advantageous.

\subsubsection{Performance on Reasoning-Intensive Tasks (BRIGHT Dataset)}

The BRIGHT dataset \cite{su2024bright} is specifically designed to evaluate models on reasoning-intensive retrieval tasks, simulating real-world scenarios where complex reasoning is required to determine document relevance. We report the Inst-L\cite{su2024bright,su2022one} model as a baseline because it is fine-tuned with task-specific instructions, making it particularly well-suited for tasks requiring deeper reasoning. Table~\ref{tab:bright_res} shows the NDCG@5 results on the BRIGHT dataset. In the Earth Science domain, our R2R model achieves an NDCG@5 of \textbf{31.0\%}, closely approaching the teacher model's \textbf{37.1\%}. Similarly, in Psychology, the student model attains \textbf{28.9\%}, compared to the teacher's \textbf{34.2\%} and BM25's \textbf{11.6\%}.
However, we observe that in certain domains, such as \textbf{Pony} and \textbf{AoPS}, the performance of our student and teacher models is relatively low.

\begin{table}[t]
\centering
\small
\caption{NDCG@5 results on the BRIGHT dataset.}
\begin{tabular}{lcccc}
\toprule
\textbf{Domain} & \textbf{BM25} & \textbf{Inst-L\cite{su2024bright}} & \textbf{R2R} & {\textbf{GPT-4 Teacher}} \\
\midrule
\textbf{Bio.}   & 17.2 & 21.0 & 24.3 & 28.0  \\
\textbf{Earth.} & 24.2 & 27.0 & 31.0 & 37.1 \\
\textbf{Econ.}  & 13.2 & 16.5 & 19.0 & 22.1 \\
\textbf{Psy.}   & 11.6 & 26.0 & 28.9 & 34.2 \\
\textbf{Rob.}   & 12.0 & 14.5 & 16.0 & 19.0 \\
\textbf{Stack}  & 16.0 & 18.5 & 21.0 & 24.2 \\
\textbf{Sus.}   & 12.4 & 16.2 & 18.0 & 21.9 \\
\textbf{Leet.}  & 22.1 & 25.0 & 27.6 & 31.4 \\
\textbf{Pony}   & 8.5  & 5.2  & 5.9  & 6.7  \\
\textbf{AoPS}   & 6.0  & 6.8  & 7.6  & 9.2  \\
\textbf{TheoQ.} & 9.4  & 15.3 & 18.5 & 23.0 \\
\textbf{TheoT.} & 3.5  & 14.2 & 17.0 & 21.7 \\
\midrule
Average & \textbf{13.1} & \textbf{17.5} & \textbf{19.6} & \textbf{23.2} \\
\bottomrule
\end{tabular}

\label{tab:bright_res}
\end{table}

\section{Ablation Study}

\subsection{Quality of Generated Reasons}

We assess the quality of the generated reasons using BLEU and ROUGE-L as presented in Table~\ref{tab:reasoning_generation} compared with the teacher model's reason. The baseline model here represents the model trained without the rank loss function and has the same input as ours.
 On the DL19 dataset, the student model achieves a BLEU score of \textbf{21.4\%}, compared to the baseline's \textbf{18.0\%}, and a ROUGE-L score of \textbf{36.8\%} versus \textbf{32.5\%}. Similar improvements are observed on DL20, where the student model attains BLEU and ROUGE-L scores of \textbf{24.0\%} and \textbf{38.2\%}, respectively, surpassing the baseline's \textbf{19.5\%} and \textbf{34.0\%}.


\begin{table}[ht]
\centering 

\caption{Evaluation of reason generation using BLEU and ROUGE-L scores for student models trained with and without the ranking loss function. The 'Student Model w/o Rank Loss' is trained without the rank loss function and has the same input as the 'Student Model w/ Rank Loss'.}

\begin{tabular}{lccc}
\toprule
\textbf{Dataset} & \textbf{Student Model} & \textbf{BLEU} & \textbf{ROUGE-L}  \\
\midrule
DL19 &  w/o Rank Loss & 18.0 & 32.5 \\
&  w/ Rank Loss & \textbf{21.4} & \textbf{36.8} \\ 
DL20 &  w/o Rank Loss & 19.5 & 34.0 \\
&  w/ Rank Loss & \textbf{24.0} & \textbf{38.2}  \\
Covid &  w/o Rank Loss & 16.0 & 30.0 \\
& w/ Rank Loss & \textbf{18.9} & \textbf{34.5}  \\
Touche & w/o Rank Loss & 17.5 & 33.0\\
&  w/ Rank Loss & \textbf{20.5} & \textbf{37.1} \\
News &  w/o Rank Loss & 20.0 & 35.0  \\
& w/ Rank Loss & \textbf{22.7} & \textbf{39.3} \\
\bottomrule
\end{tabular} 
\label{tab:reasoning_generation} 
\end{table}


\subsection{Impact of Reasoning Strategies}
In Table~\ref{tab:ablation}, by removing \textbf{comparison reasoning} results in a performance drop on DL19 from \textbf{75.3\%} to \textbf{74.1\%} (NDCG@5). Similarly, removing \textbf{ direct relevance reasoning} causes a performance decrease on DL20 from \textbf{72.3\%} to \textbf{72.2\%} (NDCG@5). Both reasoning strategies contribute to performance improvements.

\begin{table}[ht]
\centering

\caption{Ablation study: Impact of  direct relevance and comparison reasoning on NDCG@5/@10 for DL19 and DL20.}

\setlength{\tabcolsep}{4pt} 
\renewcommand{\arraystretch}{1.1} 
\resizebox{0.5\textwidth}{!}{
\begin{tabular}{lcc}
\toprule
\textbf{Model} & \textbf{DL19 NDCG@5/@10} & \textbf{DL20 NDCG@5/@10} \\
\midrule
 Direct relevance \& comparison & \textbf{75.3} / 73.8 & \textbf{72.3} / 71.22 \\
w/  direct relevance reasoning & 74.5 / 73.1 & 70.1 / 70.9 \\
w/ comparison reasoning & 74.1 / 72.9 & 72.2 / 70.4 \\
w/o reasoning & 73.2 / 72.1 & 70.8 / 69.8 \\
\bottomrule
\end{tabular}
}
\label{tab:ablation}
\end{table}

\subsection{Effects of Training Data Size}
\begin{figure}[ht]
\centering
\includegraphics[width=0.45\textwidth]{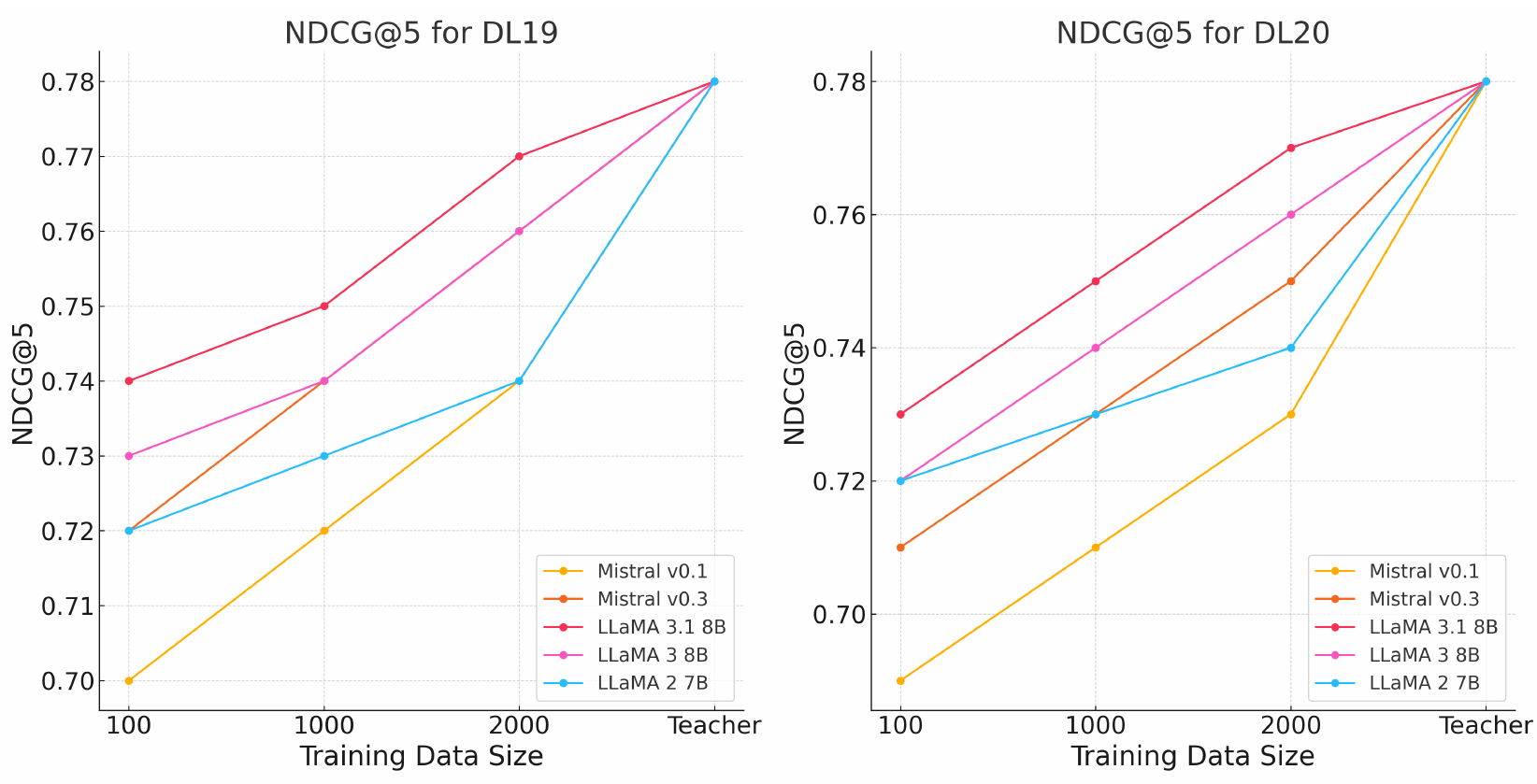}
\caption{Student model performance for different training data sizes on DL19 and DL20.}
\label{fig:training_data_size}
\end{figure}

 Figure~\ref{fig:training_data_size} shows the performance of small models trained on 100, 1,000, and 2,000 examples. Contrary to expectations, increasing data size from 1,000 to 2,000 does not consistently improve performance and sometimes slightly reduces it.  Similar trends across various small models, including Mistral and LLaMA variants, highlight that adding more data does not always guarantee better performance.

\begin{table}[t]
    \centering
    \small
    \caption{Impact of different $(\alpha, \beta, \gamma)$ combinations on model performance.
    We report the NDCG (averaged at a certain cutoff), BLEU, and ROUGE-L scores under different weighting strategies.
    The best performance is in \textbf{bold}.}
    \label{tab:hyperparam}
    \begin{tabular}{ccc|ccc}
    \toprule
    \multicolumn{3}{c|}{Hyperparameters} & \multicolumn{3}{c}{Performance} \\
    \textbf{$\alpha$} & \textbf{$\beta$} & \textbf{$\gamma$} & \textbf{NDCG} & \textbf{BLEU} & \textbf{ROUGE-L} \\
    \midrule
    0.2 & 0.2 & 0.6 & 0.78 & 0.74 & 0.72 \\
    0.3 & 0.3 & 0.5 & 0.82 & 0.76 & 0.75 \\
    0.4 & 0.4 & 0.2 & \textbf{0.86} & \textbf{0.79} & \textbf{0.78} \\
    0.5 & 0.5 & 0.1 & 0.84 & 0.77 & 0.76 \\
    0.6 & 0.6 & 0.05 & 0.80 & 0.73 & 0.74 \\
    \bottomrule
    \end{tabular}

\end{table}
\subsection{Hyperparameter Tuning.}
 We conducted a small-scale grid search over $\alpha$, $\beta$,$\gamma$ on a held-out validation set. Table~\ref{tab:hyperparam} shows the results for different weighting strategies. We find that  ($\alpha$=0.4,$\beta$=0.4,$\gamma$=0.2) yields the best trade-off among NDCG, BLEU, and ROUGE-L, and we thus adopt these settings for all subsequent experiments.

 \subsection{Case Study}
In addition to ranking improvements, the generated explanations offer interpretability benefits. For example, in several case studies, see Table \ref{tab:case_study}, we observed that the comparison reasoning often highlights subtle differences between top-ranked documents, which can assist developers in diagnosing ranking errors. Although our current evaluation of explanations relies on automated metrics (BLEU, ROUGE), a small-scale human evaluation is planned in future work to quantitatively assess readability, consistency, and practical utility of these explanations.
\begin{table}[t]
\centering
\caption{An illustrative case study showing how direct  and comparison reasoning enhance interpretability for a medical-related query. 
In this example, Document~A is ultimately deemed more relevant due to its detailed coverage of insulin’s regulatory mechanism, whereas Document~B offers a more general discussion of hormones.}
\label{tab:case_study}
\renewcommand{\arraystretch}{1.1}
\resizebox{\columnwidth}{!}{%
\begin{tabular}{p{0.95\linewidth}}
\toprule
\textbf{Query:} \emph{"How does insulin regulate blood sugar?"} \\ 
\midrule
\textbf{Document A (snippet)}\\
\quad \small\textit{``Insulin is produced by the pancreas and helps cells take in glucose from the bloodstream. By binding to insulin receptors, it initiates a signaling cascade that lowers blood glucose levels...''}\\
\textbf{Direct Reasoning:}\\
\quad \small\textit{``This document directly addresses how insulin facilitates cellular glucose uptake, which aligns precisely with the question on regulating blood sugar.''}\\
\textbf{Comparison  Reasoning vs. Document B:}\\
\quad \small\textit{``Compared to Document B, Document A provides more explicit details on insulin’s binding mechanism and the biochemical pathway, making it more relevant for understanding exactly how insulin regulates glucose.''}\\
\midrule
\textbf{Document B (snippet)}\\
\quad \small\textit{``Hormones like insulin and glucagon maintain metabolic balance. Insulin generally lowers blood sugar levels, while glucagon raises them, ensuring homeostasis...''}\\
\textbf{Direct Reasoning:}\\
\quad \small\textit{``This document briefly mentions insulin’s role in lowering blood sugar, but primarily focuses on the broader hormone interplay.''}\\
\textbf{Comparison  Reasoning vs. Document A:}\\
\quad \small\textit{``Although Document B is relevant, it emphasizes multiple hormones and offers less specific detail on the mechanism of insulin. Document A is therefore more closely aligned with how insulin regulates blood sugar.''}\\
\bottomrule
\end{tabular}%
}
\end{table}

\section{Discussion and Conclusion}
\label{sec:discussion}
Our experimental results demonstrate that \textbf{Reason-to-Rank (R2R)} effectively enhances document reranking by integrating both direct relevance and comparative reasoning. Compared to relying solely on a large teacher model (e.g., GPT-4), our distilled student models offer \textit{competitive performance} while drastically reducing computational cost, making the framework suitable for real-world, high-throughput settings. The statistically significant gains confirm that \emph{incorporating explicit reasoning structures}---both pointwise and pairwise---meaningfully contribute to improved ranking accuracy.

A natural concern is whether the observed performance improvements stem merely from generating longer output text rather than the intrinsic value of direct and comparison reasoning. We argue that the benefit of our approach lies in the structured content of the explanations: by explicitly contrasting documents (comparison reasoning) and detailing query-document alignment (direct reasoning), the model captures fine-grained distinctions that standard ranking scores might overlook. Although we have not conducted a controlled experiment to equalize text length, preliminary analyses indicate that even when constraining the output to a fixed token budget, models incorporating both reasoning types consistently outperform those without. Future work will include a more rigorous ablation study and a human evaluation to further assess the usefulness of the generated rationales in aiding system diagnosis and debugging.

Notably, the proposed student models not only output document rankings but also produce coherent and relevant justifications for each document’s position. This dual capacity to \emph{rank} and \emph{explain} suggests broader applicability beyond traditional ad-hoc retrieval. For example, in \emph{recommendation systems}, R2R could prioritize items while elucidating the rationale (e.g., product features or user context), thereby increasing user trust and satisfaction. Similarly, in \emph{question answering} or \emph{summarization} tasks, generating reasoning alongside answers may boost transparency and user acceptance~\cite{jiang2024trisum}. However, systematically evaluating these broader domains remains as future work to verify the generalizability of our approach~\cite{mcelfresh2022generalizability, rahmani2022experiments}.

Overall, the synergy between direct relevance and comparative reasoning leads to both practical efficiency and rich interpretability. Our findings point to an emerging trend in IR and NLP systems where explicit reasoning improves performance metrics and contributes to more explainable and user-friendly outcomes.

\label{sec:conclusion}
In this paper, we introduced \textbf{Reason-to-Rank (R2R)}, a novel framework that unifies \emph{direct relevance} and \emph{comparison} reasoning for document reranking. We leveraged large language models (LLMs) as teacher models to generate high-quality explanations and rankings, then distilled both the ranking expertise and textual justifications into smaller student models. Experimental results across multiple benchmarks illustrate that R2R not only achieves strong reranking effectiveness but also delivers interpretable rationales for why one document is ranked above another. Our study underscores the value of distilling explicit reasoning strategies from large LLMs and highlights the benefits of merging pointwise and pairwise perspectives.

\section{Limitations and Future Work}
\label{sec:limitations}
Despite the encouraging results, there are several limitations to consider:
 \textbf{Dependence on large LLMs for teacher data.} Our approach relies on powerful teacher models (e.g., GPT-4) to generate high-quality explanations. Obtaining robust supervision signals could be challenging in resource-constrained environments or specialized domains where such models are not readily available. Also, using GPT-4 can be expensive (approx.\$0.2 per query with 4,000+ tokens). While knowledge distillation substantially reduces cost at inference time, the distilled models are still large and thus not trivial for real-time production in specific scenarios.
\textbf{Prompt quality and model biases.} The effectiveness of the distilled reasoning hinges on carefully designed prompts. If prompts fail to capture key aspects of the query or document, the generated explanations might be shallow or incorrect. Additionally, teacher models can introduce biases, which might propagate into student models during distillation.

 \textbf{Broader applicability beyond reranking.} While we have discussed possible extensions to recommendation, QA, or summarization tasks, the actual performance and adaptation strategy in these domains remain largely unexplored. Thorough empirical studies are required to confirm that the proposed dual reasoning approach generalizes effectively. \textbf{Lack of Human Evaluation for Explanations} We only measured textual explanations with BLEU/ROUGE against the teacher outputs, which themselves are automatically generated. This does not guarantee that the explanations are genuinely helpful or correct from a human perspective. \textbf{Generalization beyond Document Ranking.} Though we focus on ad-hoc retrieval, the idea of distilling direct and comparative reasoning could benefit other tasks like question answering or recommender systems. More thorough domain-specific experiments are necessary to confirm the general applicability of our approach.\textbf{Confounding Factors of Explanation Length.}While our ablation controls for token length, further studies are needed to rigorously isolate how much of the ranking improvement stems from the reasoning structure vs.\ extended context. Understanding how best to prompt for \emph{concise} but \emph{effective} rationales remains an open question.

In future work, we plan to investigate more efficient teacher–student paradigms (e.g., smaller teacher models, automated prompt generation) and conduct domain-specific evaluations to further validate the versatility of R2R. We believe these efforts can refine the balance between interpretability and computational cost, extending the framework’s applicability across diverse IR and NLP scenarios.
\section{Appendix}

\subsection{Student Training Parameters}

\label{sec:appendixA}

\begin{table}[htbp]
\centering
\caption{Training configuration hyperparameters.}
\renewcommand{\arraystretch}{0.9} 
\small
\begin{tabular}{lc}
\toprule
\textbf{Hyperparameter}            & \textbf{Value}\\ \hline
low\_cpu\_fsdp                     & True                         Interpretability  \\ 
run\_validation                    & True                                           \\ 
batch\_size\_training              & 1                                              \\ 
context\_length                    & 1024                                            \\ 
gradient\_accumulation\_steps      & 1                                              \\ 
gradient\_clipping                 & True                                           \\ 
gradient\_clipping\_threshold      & 1.0                                            \\ 
num\_epochs                        & 3                                              \\ 
max\_train\_step                   & 0                                              \\ 
max\_eval\_step                    & 0                                              \\
learning rate                      & 1e-4                                           \\
weight\_decay                      & 0.0                                            \\ 
seed                               & 42                                             \\ 
use\_fp16                          & True                                           \\ 
use\_peft                          & True                                           \\ 
freeze\_layers                     & False                                          \\ 
\bottomrule
\end{tabular}
\end{table}
\subsection{API Cost Analysis}
We calculated the average number of tokens, the 
number of API requests, and the estimated cost per 
query for different operations involving the teacher 
model, see in Table \ref{tab:cost}. The costs are based on the pricing provided by the API service at the time of our experiments.

\begin{table}[ht]
\centering
\caption{Average token usage and cost per query using GPT-4 for different reasoning tasks.}
\small
\begin{tabular}{lccc}
\toprule
\textbf{API} & \textbf{Instruction} & \textbf{Tokens}  & \textbf{Cost (\$USD)} \\
\midrule
GPT-4 &  Direct relevance  & 3650  & 0.134 \\
GPT-4 & Comparison  & 4050  & 0.158 \\
GPT-4 &  Direct relevance \& Comparison  & 4650  & 0.194 \\
\bottomrule
\end{tabular}
\label{tab:cost}
\end{table}

\bibliographystyle{ACM-Reference-Format}
\balance
\bibliography{sample-base}

\appendix

\end{document}